\documentclass[letterpaper]{article} 
\usepackage{aaai25}  
\usepackage{times}  
\usepackage{helvet}  
\usepackage{courier}  
\usepackage[hyphens]{url}  
\usepackage{graphicx} 
\usepackage{natbib}  
\usepackage{xspace} 
\usepackage{ragged2e} 
\usepackage{booktabs,makecell, multirow, tabularx}
\usepackage{caption} 
\usepackage[table]{xcolor}
\usepackage{enumitem}

\usepackage{algorithm}
\usepackage{algorithmic}
\usepackage{amsmath}
\usepackage{bbold}
\usepackage{amsfonts} 
\usepackage{amssymb}
\usepackage{pifont}
\usepackage{makecell}
\usepackage{bbding}

\usepackage{scalerel,graphicx,xparse}
\newcommand{\onedot}{.\xspace}

\def\ie{\emph{i.e}\onedot}

\makeatother

\newcommand{\myparagraph}[1]{\vspace{1ex}\noindent\textbf{#1.}\hspace{1em}}

\newcommand{\amg}{\textsf{AMG}\xspace}
\newcommand{\our}{\textsf{MGCA}\xspace}

\def\ie{\emph{i.e}\onedot}

\usepackage{newfloat}
\usepackage{listings}
\DeclareCaptionStyle{ruled}{labelfont=normalfont,labelsep=colon,strut=off} 
\floatstyle{ruled}
\newfloat{listing}{tb}{lst}{}
\floatname{listing}{Listing}

\newcommand{\sota}{\textsf{state-of-the-art}\xspace}
\title{\emph{Each Fake News is Fake in its Own Way:} An Attribution Multi-Granularity Benchmark for Multimodal Fake News Detection}
\author{
    Hao Guo\equalcontrib\textsuperscript{\rm 1},
    Zihan Ma\equalcontrib\textsuperscript{\rm 2,3,4},\\
    Zhi Zeng\textsuperscript{\rm 2,3,4},
    Minnan Luo\textsuperscript{\rm 2,3,4},
    Weixin Zeng\textsuperscript{\rm 1},
    Jiuyang Tang\textsuperscript{\rm 1},
    Xiang Zhao\thanks{Corresponding Author.}\textsuperscript{\rm 1},
}
\affiliations{
    \textsuperscript{\rm 1}Laboratory for Big Data and Decision, Nation University of Defence Technology,\\
    \textsuperscript{\rm 2}School of Computer Science and Technology, Xi’an Jiaotong University,\\
    \textsuperscript{\rm 3}Ministry of Education Key Laboratory of Intelligent Networks and Network Security,
    Xi’an Jiaotong University, \\
    \textsuperscript{\rm 4}Shaanxi Province Key Laboratory of Big Data Knowledge Engineering, Xi’an Jiaotong University.
    
    \texttt{\{guo\_hao,jiuyang\_tang,xiangzhao\}@nudt.edu.cn},
    \texttt{\{mazihan880,zhizeng\}@stu.xjtu.edu.cn}
    \\ \texttt{minnluo@xjtu.edu.cn,dexterdervish@foxmail.com}
}

\usepackage{bibentry}

\begin{document}

\maketitle

\begin{abstract}
Social platforms, while facilitating access to information, have also become saturated with a plethora of fake news, resulting in negative consequences. Automatic multimodal fake news detection is a worthwhile pursuit. Existing multimodal fake news datasets only provide binary labels of real or fake. However, real news is alike, while each fake news is fake in its own way. These datasets fail to reflect the mixed nature of various types of multimodal fake news. To bridge the gap, we construct an attributing multi-granularity multimodal fake news detection dataset \amg, revealing the inherent fake pattern. Furthermore, we propose a multi-granularity clue alignment model \our to achieve multimodal fake news detection and attribution. Experimental results demonstrate that \amg is a challenging dataset, and its attribution setting opens up new avenues for future research.
\end{abstract}

\begin{links}
    \link{Code and Datasets}{https://github.com/mazihan880/AMG-An-Attributing-Multi-modal-Fake-News-Dataset.}
    \link{Extended version}{https://aaai.org/example/extended-version}
\end{links}

\section{Introduction}
Fake news is false or misleading information presented as news~\cite{rubin2016fake,molina2021fake}.
Social media platforms are inundated with fake news, exerting a significant impact on public health, governance, and societal equilibrium~\cite{zannettou2019web, allcott2017social,apuke2021fake}. 
In recent years, the media-rich nature of these platforms has led to a gradual shift in the type of information shared by the public, encompassing not only textual content but also a plethora of visual elements such as images and videos. Because of the  ``Multimedia Effect''~\cite{mayer2002multimedia}, multimedia content such as images and videos exerts a heightened allure on individuals~\cite{jamet2008attention,38mayer2014incorporating}. 
Furthermore, visual content is commonly utilized as substantiating evidence within storytelling, thus augmenting the credibility of news narratives.
Regrettably, fake news publishers have adeptly utilized this opportunity to captivate attention and enhance credibility, leading to an evolution towards multimodal formats~\cite{cao2020exploring}. 
The task of multimodal fake news detection has grown progressively intricate, which is the focal point of our research.

\begin{figure}[t] 
    \centering
     \includegraphics[width=1.02\linewidth]{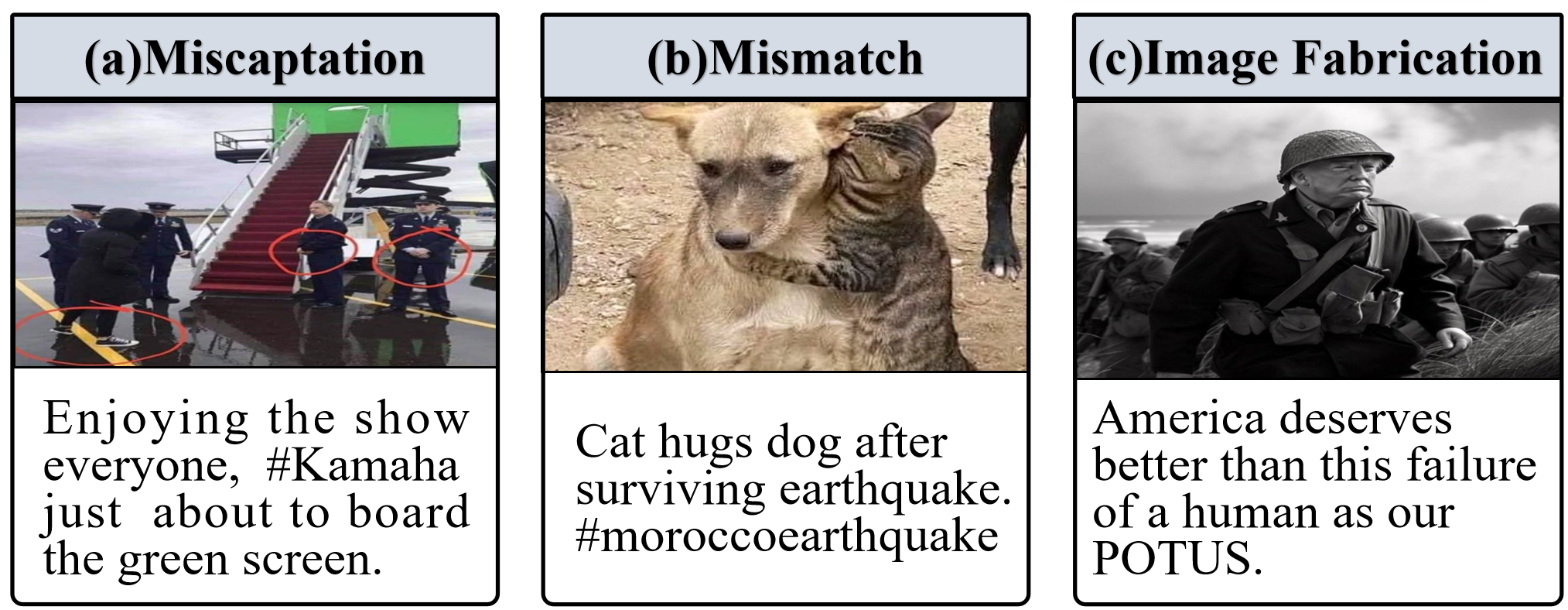}
    \caption{Various types of multimodal fake news in Twitter. "Miscaption" means that the caption of the image does not match the text. "Mismatch" indicates the image is related to the text, but from previous similar event. "Image Fabrication" indicates that the image comes from deepfake technology but is not stated.}
    \label{multitype}
\end{figure}

In contrast to news that relies solely on textual content, multimodal fake news encompasses visual, textual, and cross-modal correlation, allowing fabricators to craft deceptive narratives from multiple perspectives. We have observed that \emph{real news is alike, each fake news is fake in its own way}. In popular social platforms Twitter, multimodal fake news manifests in various distinct types\footnote{Disclaimer. All examples of fake news in this paper are for illustrative purposes only and do not depict real incidents or accurate information. Any resemblance to actual persons or events is purely coincidental.}, as depicted in Figure~\ref{multitype}. 
However, existing multimodal fake news detection methods typically focus on only one type. 
\textbf{First}, some methods incorporate visual-textual consistency features into the basis for detection~\cite{safe,63qi2021improving}, which aim to capture the correlation between the textual and visual content. The methods focus on detecting types like Figure~\ref{multitype}(a), where the key person ``Kamala'' does not appear in attached image, while such methods overlook the temporal information. 
\textbf{Second}, a plethora of fake news utilizes images from other times and places for the latest trending events, as depicted in Figure~\ref{multitype}(b): a picture of Turkey earthquake in Feb. 2023 is used to describe the Morocco earthquake in Sep. 2023, creating a strong association between the image and the accompanying text. 
\textbf{Third}, manipulated images directly impact the authenticity of news~\cite{59jin2016novel}. 
Current methods exploit the frequency domain~\cite{wu2021multimodalmcan} and pixel domain~\cite{qi2019exploitingmvnn} features of images to detect multimodal fake news, which tend to fall short due to the proliferation of Artificial Intelligence Generated Content (AIGC)~\cite{huang2023implicit,rombach2022high} that poses a significant challenge in combating deepfake images~\cite{xu2023combating,shao2023detecting}. As shown in Figure~\ref{multitype}(c), the image of ``Trump while serving in the military'' is deepfake.

Despite of the various types of  multimodal fake news, existing detection solutions have not fully considered the scenario where multiple types of fake news coexist and ignored the time consistency cross the image and text. Besides, most models can only output authenticity scores, which are compared with the authenticity labels in the datasets. 
However, the labels in the datasets are derived directly from fact-checking agencies, with the majority consisting of binary labels indicating only real or fake~\cite{nan2021mdfend, 65boididou2015verifying}, without providing fine-grained attribution labels that reveal the error patterns in multimodal fake news.
Inspired by the idea of attributing unanswerable questions in the question answering domain~\cite{rajpurkar2018know,liao2022ptau}, if we can attribute the types of multimodal fake news while detecting its authenticity, the credibility of the detection model will be further enhanced.  
Although a very recent study explores deception patterns in multimodal fake news~\cite{Dong_He_Wang_Jin_Ge_Yang_Jin_2024}, there still lack benchmarks and effective solutions for attributing multimodal fake news.



To surmount the constraints, we develop the first dataset for \underline{a}rrributing \underline{m}ultimodal fake news with multi-\underline{g}ranularity, namely \amg. To build the dataset, we collect fake news from multiple platforms. Then attribution rules are designed, and expert annotation is performed based on the rules and ruling articles from fact-checking websites. Finally, a three-fold cross-validation is conducted to achieve fine-grained attribution of fake news.

Furthermore, we propose a multimodal fake news detection and attribution model based on \underline{m}ulti-\underline{g}ranular \underline{c}lues \underline{a}lignment, namely \our. 
It extracts multi-view features from both visual and textual contents and incorporates consistency modeling of multi-granular clues to aid in authenticity detection and attribution. 
Extensive experimental results and analyses provide evidence for the increased challenge posed by our proposed dataset. 
Overall, our contributions are three-fold:

    (1) To the best of our knowledge, we are among the first to elicit the notion and motivate the challenges of multi-granularity multimodal fake news attribution.

    (2) Our proposed \amg is a first fine-grained attribution of multimodal fake news based on the causes of fake, attributing them to image fabrication, non-evidential image, entity inconsistency, event inconsistency and time inconsistency.

    (3) We propose \our, a strong baseline to handle multimodal fake news detection and attribution, the performance of which is demonstrated by comprehensive experiments on \amg.      

\begin{table*}
{
    \centering
    \setlength{\tabcolsep}{5pt}
    \renewcommand{\arraystretch}{0.9}
    \caption{Compilation of multimodal fake news datasets. \#Post represents the number of multimodal news piece. \#Image represents the number of unique image. MR$^2$ has both Twitter and Weibo datasets.}
    \label{alldataset}
    \scalebox{0.9}{
    \begin{tabular}{c c c c c c c c c}
    \Xhline{1.5px}
        Datasets & \thead{Time period} & Class & \#Post & \#Image & Source & \thead{Attribution} & Domain & \thead{Temporal Info}\\ \hline
        \thead{Weibo21~\cite{nan2021mdfend}} & 2014-2021  & 2 & 9,128 & - & Weibo & \ding{53} & variety & \ding{53}  \\ 
        \thead{Weibo~\cite{jin2017multimodal}} & 2012-2016  & 2 & 9,528 & 9,528 & Weibo & \ding{53} & variety & \ding{51} \\ 
        \thead{PolitiFact~\cite{64shu2020fakenewsnet}} & -2020 & 2 & 359 & 359 & Twitter & \ding{53} & politics & \ding{51}\\ 
        \thead{GossipCop~\cite{64shu2020fakenewsnet}} & -2020  & 2 & 10,010 & 10,010 & Twitter & \ding{53} & gossip & \ding{51} \\ 
        \thead{Twitter~\cite{65boididou2015verifying}} & -2014 & 2 & 13.924 & 514 & Twitter & \ding{53} & 11 events & \ding{51}\\ 
        \thead{ReCOVery~\cite{66zhou2020recovery}} & -2020 & 2 & 2,017 & 2,017 & Twitter & \ding{53} & covid-19 & \ding{51} \\ 
        \thead{Pheme~\cite{67zubiaga2017exploiting}} & 2014-2015 & 2 & 5,802 & 3,670 & Twitter & \ding{53} & 5 events & \ding{51} \\ 
        \thead{Fakeddit~\cite{nakamura2020fakeddit}} &  2008-2019 & 2/3/6 & 682,996 & 682,996 &  Reddit & \ding{53} & variety & \ding{51}\\
        \thead{MR$^2$~\cite{68hu2023mr2}} & -2022 & 3 & \thead{ 7,724 \\ 6,976}  & \thead{7,724 \\ 6,976} & \thead{Twitter \\ Weibo} & \ding{53} & variety & \ding{53}\\ 
        \hline
        \amg & 2016-2024 &  2/6 &  5,022 & 5,022 & \thead{Ins/Twitter \\ Facebook} & \ding{51} & variety & \ding{51}\\ \Xhline{1.5px}
    \end{tabular}}}
\end{table*}

\section{Dataset Construction}
\label{data}
\amg, as the pioneering dataset for multimodal fake news detection and attribution, encompasses posts originating from diverse social platforms. In this section, the data collection, data processing and annotation, and the collation and analysis of \amg will be described in detail.

\subsection{Data Collection}
\myparagraph{Fake News Collection}
For gathering fake news, we intend to utilize existing fact-checking websites as initial sources of news. The ruling articles found on these websites can assist in fine-grained type annotation. Among them, Snopes\footnote{https://www.snopes.com} and CHECKYOURFACT\footnote{https://checkyourfact.com} are widely recognized websites that verify and expose fake news. 
Professionals, including journalists, gather pertinent evidence and engage in evidence-based reasoning to formulate ruling articles, providing judgments on the authenticity of news.
Instead of crawling short claims from the titles of fact-checking websites~\cite{yao2023end}, we crawl the original posts associated with claims from various platforms, primarily including Instagram, Facebook, Twitter, TikTok, and YouTube, which aligns more closely with the reality of fake news on social platforms. Among them, we focus on Instagram, Facebook, and Twitter as the main sources of these posts. 

\myparagraph{Real News Collection}
Initially, we crawl the verified true news from the same fact-checking websites. However, the quantity obtained is quite limited (only 126). 
Besides, to mitigate inherent biases between real and fake news~\cite{zhu2022generalizing}, it is essential to establish a relatedness between real news and the corresponding fake news.
Therefore, we compensate for the shortage of real news by the following steps.
Firstly,  we employ the pre-trained Large Language Model Vicuna~\cite{zheng2023judging} as our entity extraction tool.
Then, based on the distribution proportion of fake news on social platforms, we crawl real news associated with these entities from authoritative and neutral media accounts\footnote{https://www.allsides.com/unbiased-balanced-news}, such as Reuters and NewsNation. 

Due to an insufficient number of retrieved related real news, we have randomly selected a certain quantity of news articles from the aforementioned official account's archive to supplement the dataset. These news span from 2016 to 2023, aligning with the temporal scope of the fake news. 
To maintain a similar ratio to the previous dataset, we set the quantity of genuine news to be 1.5 times that of fake news.

\subsection{Data Processing and Annotation}
\myparagraph{Filtering}
In order to construct a multimodal fake news dataset, our first step involves filtering news articles based on the presence of relevant multimodal news. Both images and videos are included within the scope of our dataset.
Moving forward, we utilize visual content similarity to eliminate news articles with high resemblance to one another, thus preserving the diversity among each piece and safeguarding against potential data leakage.

\begin{figure}[t] 
    \centering
     \includegraphics[width=1.05\linewidth]{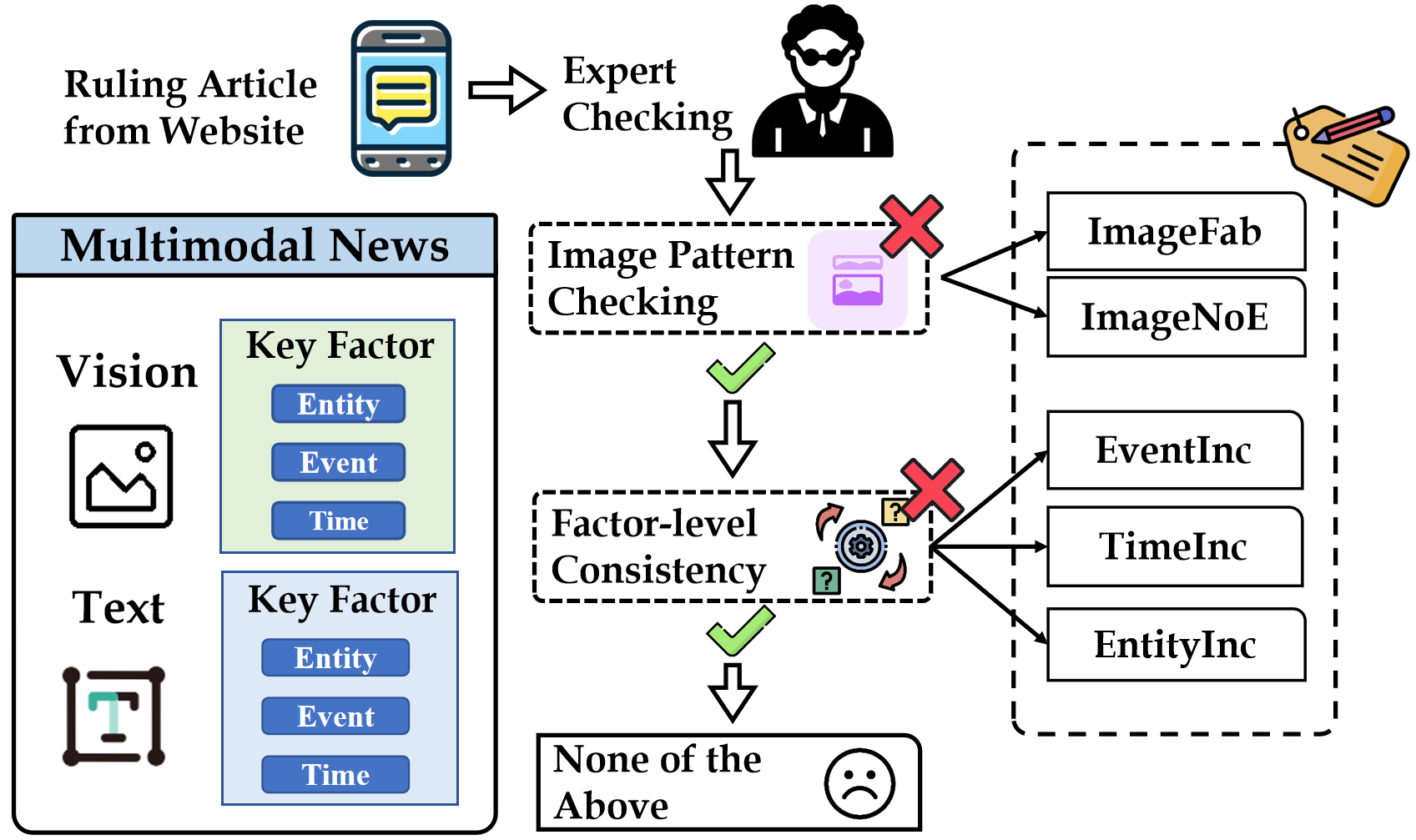}
    \caption{The process of multimodal fake news attribution.}
    \label{attProcess}
\end{figure}

\myparagraph{Expert Annotation}
Diverging from our previous approach of directly crawling websites with authenticity labels, we have embarked on a more detailed annotation process for news articles, based on these labels and verified articles. Our annotation work is carried out by a team of experts who possess relevant domain knowledge. A comprehensive annotation guideline has been developed, along with specialized training for the annotators. The team consists of a total of 17 individuals (Details in Supplementary).

\myparagraph{Annotation Process} Binary labels indicating the truthfulness of news can be easily obtained from verification websites. For fake news, we have meticulously designed each step of the attribution process, as shown in Figure~\ref{attProcess}.  Ruling articles serve as our basis for judgment. Firstly, we perform image pattern checking on the image itself to identify any signs of fabrication or non-evidential content. Secondly, we examine the consistency between the image and the accompanying text across various key factors, attributing entity, event and time inconsistency. 
It is also possible that some instances do not belong to any of the above categories.

\begin{figure}[t] 
    \centering
     \includegraphics[width=1\linewidth]{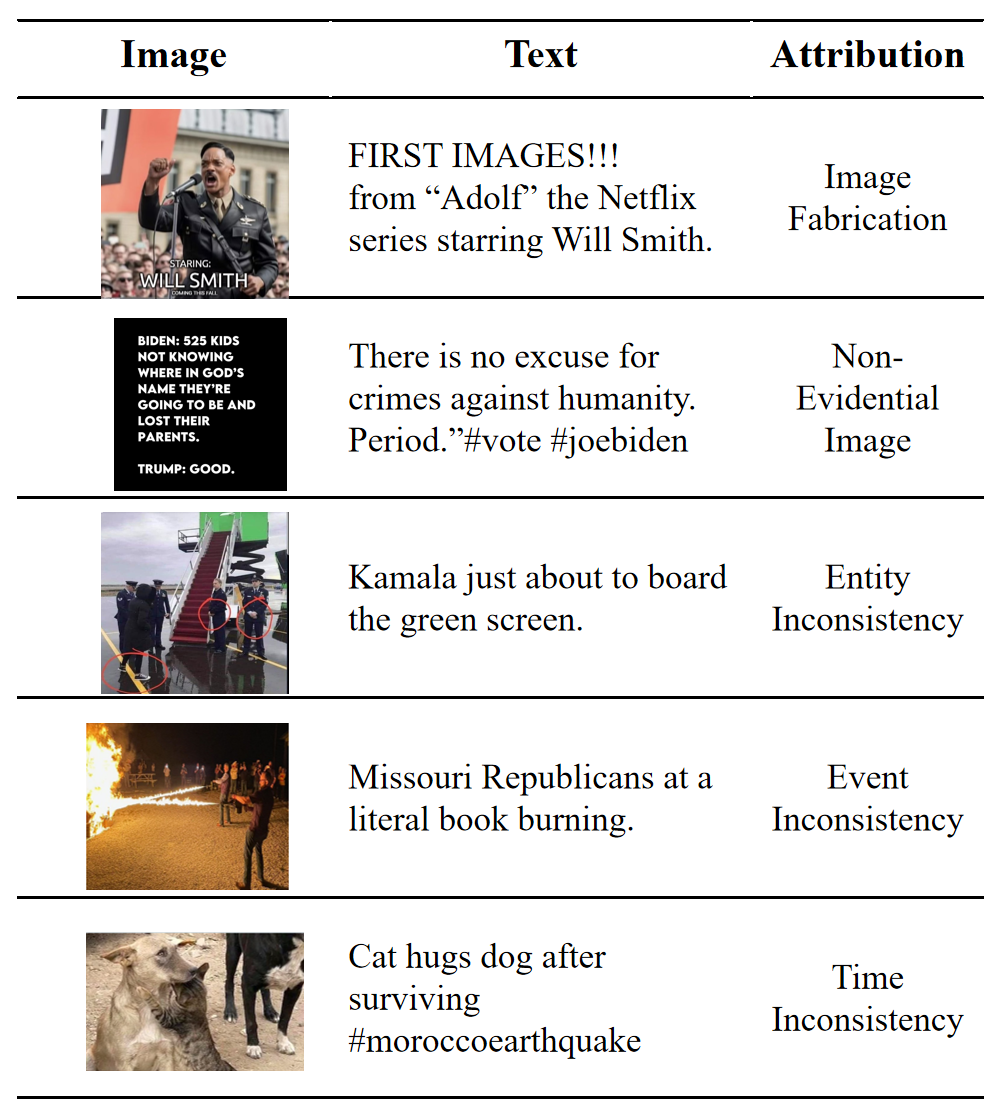}
    \caption{Examples of various attributions.}
    \label{attribution5}
\end{figure}

\myparagraph{Attribution Foundation} 
The specific explanations and theoretical foundations for each attribution type are as follows (See examples in Figure~\ref{attribution5}):

\emph{Image Fabrication (ImageFab):} The authenticity of an image is questionable. This can encompass the application of cutting-edge deepfake techniques as well as simpler forms of manipulation such as image splicing or PS. 
Furthermore, it also includes the simulation of images imitating official websites or tweets, representing a unique circumstance within the realm of image forgery.
Previous research~\cite{wu2021multimodalmcan,xue2021detectingMCNN} has already highlighted the use of the authenticity of the image for detection, while ~\cite{shao2023detecting} established a dataset for detecting AIGC-based
fake images. So image fabrication is one typical fake attribution of multimodal news.

\emph{Non-Evidential Image (ImageNoE)} refers to cases where the image consists of textual information that cannot provide evidence or proof for news content. A notable characteristic of real news is that its images provide support for the accompanying text, such as on-site photos of breaking events. On the other hand, images that solely consist of text are a common image pattern found in multimodal fake news.


\emph{Entity Inconsistency (EntityInc)} refers to a phenomenon where there is a discrepancy between the key entities depicted in the textual and visual modalities. In other words, there is a lack of alignment or coherence between the entities described in the text and those visually represented, which has been validated as an effective clue in previous study~\cite{qi2021improving, li2021entity}.

\emph{Event Inconsistency (EventInc):} Despite the presence of associated entities in both text and image, there is a event-level discrepancy. News always describes events, the alignment of textual and visual events serves as a vital criterion for assessing the authenticity of news~\cite{wei2022modality,wang2018eann}. Within this category, the images themselves are not forged, the inconsistency often arises from excessive inference and misrepresentation in the written text for attached image. 

\emph{Time Inconsistency (TimeInc)} maintains consistency at the entity or event level, a disparity arises at the temporal information level. 
It refers to the practice of using unaltered images or videos that depict past events, like natural disasters or gatherings, but falsely presenting them as recent events.
Most of out-of-context misinformation~\cite{Luo_Darrell_Rohrbach_2021,Abdelnabi_Hasan_Fritz_2022,qi2024sniffer} or image-repurposing~\cite{jaiswal2019aird,Sabir_AbdAlmageed_Wu_Natarajan_2018} can be attributed to TimeInc. 

During the labeling process, we acknowledge that there may be special cases that do not fit into our predefined categories. To account for such situations, we include the label "\textbf{None of the Above}" to accommodate those instances.
The specific examples that fall outside our attribution categories, as well as the analysis of this particular category, can be found in Supplementary.

\myparagraph{Cross Validation and Discussion}
Each fake news is assigned to three annotators, and the final attribution is determined through a majority vote following ~\cite{feng2022twibot}. Furthermore, controversial cases undergo discussion and then secondary round of annotation.

\subsection{Data Collation and Analysis}
After integrating the collected news, we filter out fake news that does not fall under our attribution types. And the quantities for each attribution type are as follows: 434, 295, 133, 667, 475.
The number of multimodal fake news from Instagram, Twitter, and Facebook are 142, 558, and 1,304, respectively. 
In addition, the final number of real news is set to approximately 1.5 times the number of fake news. The counts for real news and fake news are 3,018 and 2,004. More statistic
are listed in Supplementary.

\myparagraph{Train/Val/Test Split}
We split the whole dataset into training (Train), validation (Val), and test (Test) sets with the number of 3,532, 517 and 973, respectively. The percentage is nearly 7:1:2. Furthermore, we maintain consistent proportions within each subcategory during the dataset's split. 


\myparagraph{Rationality of our attribution rules} 
Upon analyzing the final statistics,  we make an exciting observation: the samples that fall outside our attribution categories account for only around 3\% of the total dataset, comprising approximately 60 instances.
This observation suggests that our classification rules effectively cover almost all cases of fake news, thereby confirming the soundness of our attribution guidelines.

\myparagraph{Legal and Ethical} 
Firstly, we adhere to the data scraping rules of each platform. 
Additionally, all annotators underwent rigorous training and were well-versed in data privacy and security regulations. 
During the annotation process, the annotators conducted a screening, selecting only posts related to public figures or public events, without involving ordinary users.  Furthermore, any associated personal user information was anonymized, including id and name.
We also took measures during data processing and training to prevent any leakage of user privacy((Details in Supplementary). 
All collected data is stored on secure servers, with access restricted to our research team members only.

\myparagraph{The strength of \amg} 
(1) Up-to-date and Temporal-inclusive.  The fake news in \amg originate from the period between 2020 and 2024, with a small portion encompassing February 2024. \amg includes the publication timestamps of news posts, whereas MR$^2$~\cite{hu2023learn} does not.
(2) Multiple platforms. \amg is platform-agnostic which incorporating content from the three major mainstream social platforms. 
(3) Multiple domains. Upon a simple aggregation, we find that it encompasses multiple fields such as healthcare, elections, military, entertainment, and more.
(4) Multi-granularity attribution labels. Different from Fakeddit~\cite{nakamura2020fakeddit}, the fine-grained labels of \amg reveals the attribution for fake pattern.

\section{Methodology}
\label{meth}

The section primarily discusses our proposed detection and attribution model.
(Preliminary in Supplementary) 

\begin{figure}[t]
    \centering
        \includegraphics[width=1.04\linewidth]{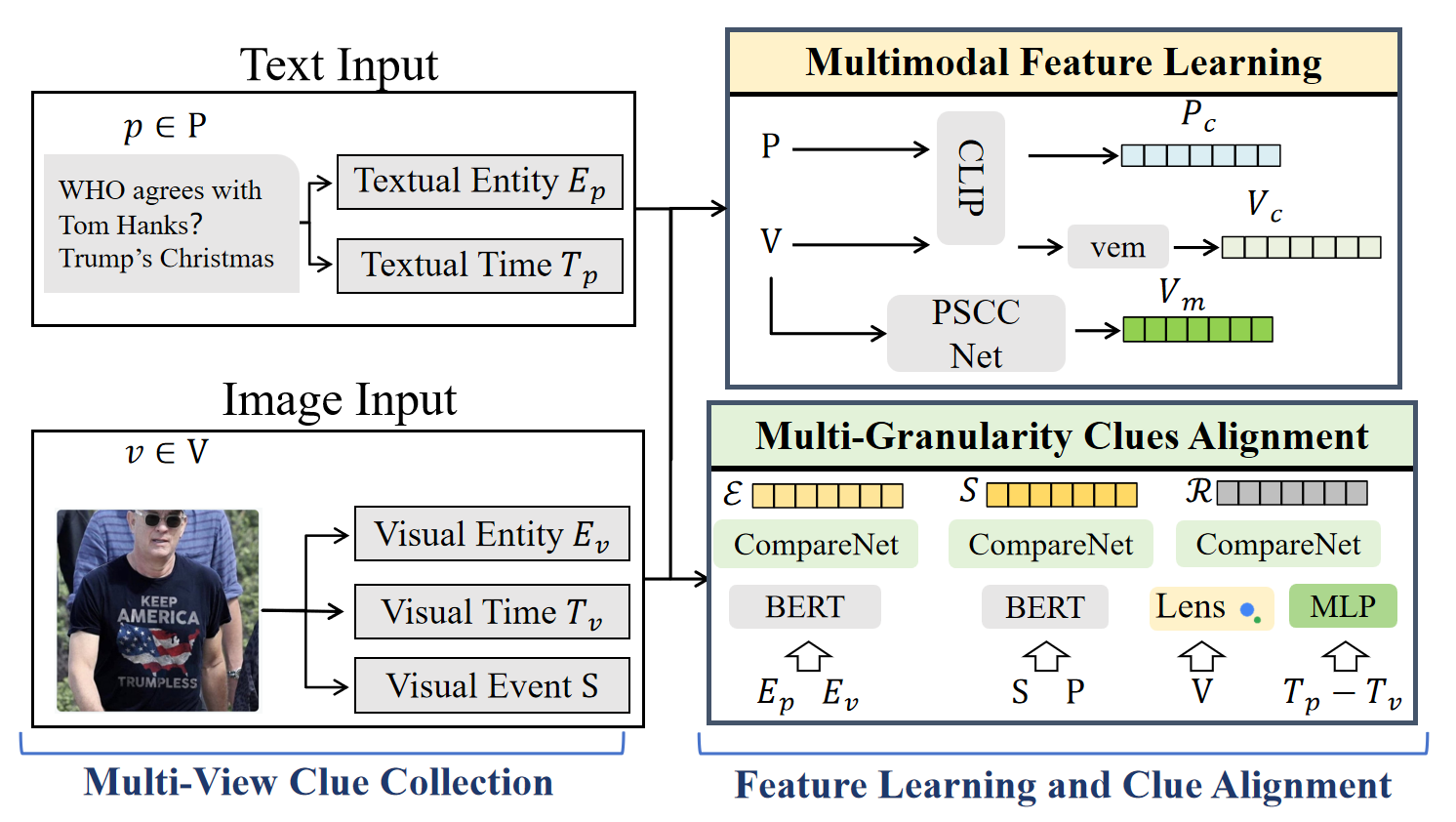}
    \caption{Model outline of \our.}
    \label{frame}
\end{figure}

\myparagraph{Model Outline} 
As depicted in Figure~\ref{frame}, \our first gathers multi-perspective clues from both images and text. Next, it performs multimodal feature learning and aligns the collected clues. Finally, it integrates the extracted features to conduct inference of detection and attribution.

\subsection{Multi-View Clue Collection}
We extract the multi-view clues from both the textual input and visual input, which includes time, entity and event. 

\myparagraph{Textual Entity}
Due to the narrative style typically present in news articles, which includes crucial named entities such as characters and locations, the association between these key entities can be instrumental in detecting fake news~\cite{qi2021improving}. To enhance this process, we employ the pre-trained Large Language Model Vicuna~\cite{zheng2023judging}. By designing prompt templates and utilizing the capabilities of the large-scale model's In-context Learning~\cite{mann2020language,xie2021explanation}, we incorporate examples of entity extraction within these templates, guiding the process. We denote the entity in the text as $E_p$.

\myparagraph{Visual Entity}
Corresponding to the textual content, certain news articles also contain valuable visual entities within their visual content. For the extraction of visual entities, we utilize Baiduan APIs\footnote{https://ai.baidu.com/tech/imagerecognition/general} that specializes in extracting three types of entities: individuals, landmarks, and organizations. We denote this extraction of visual entities as $E_v$.

\myparagraph{Textual Time}
Temporal mismatch is a significant type of multimodal fake news. In this article, we consider the temporal information of news as a crucial factor in determining its authenticity.
Firstly, we extract the time label of the news, denoted as $t_1$. As news articles often describe past events, we also extract the mentioned time, $t_2$, from the textual content. We then select the earlier time as the temporal reference for the text, which we refer to as $T_p=min\{t_1,t_2\}$.

\myparagraph{Visual Time}
Retrieving the original publication time of an image, along with its relevant content, can be helpful in identifying temporal inconsistencies in multimodal fake news.
We employ GoogleLens\footnote{https://lens.google.com/} for performing reverse image searches. By conducting such searches, we obtain the earliest corresponding time $T_v$ and title $R$ of the related image.

\myparagraph{Image Event}
In addition to visual entities, we believe that the event present in images is also a valuable auxiliary clue. We utilize multimodal large language model LLaVA~\cite{liu2023llava} for extracting image events denoted as $S$ (Conducting details in Supplementary).

\subsection{Multimodal Feature Learning}
To enhance the consistency representation, we employ CLIP~\cite{radford2021learning} to extract features $P_c$ and $V_c$ from the total of news text $P$ and news image $V$. 
To obtain the rich semantic clue representations, we exploit utilize BERT\cite{kenton2019bert} to acquire $C_s$, $C_r$, $C_p$ and $C_v$ from event clues $S$. Also, we use Bert to encode entity clues $E_p$ , $E_v$ and the retrieval clues $R$.

To ensure mathematical distribution consistency, we also utilize BERT to obtain the semantic representation $P_b$ of the news text. 
As for the timeline, we calculate the temporal gap $T_g$ between the images and the text, denoted as $T_g = (T_P-T_V)$ to characterize the temporal inconsistency.

To detect the manipulated image, we employ the effective manipulation detection network PSCC-NET~\cite{liu2022pscc} for detecting image manipulation. Specifically, by freezing the feature extraction layer of PSCC-NET, we obtain the manipulation features $V_m$ for the news images.

\subsection{Multi-Granularity Clues Alignment}
To detect the entity-level and event-level consistency between news image and text, we utilize a Compare-Net\cite{shen2018improved} to obtain consistency features $\mathcal{E}$ and $\mathcal{S}$,\ie,
\begin{equation}
\begin{aligned}
    \mathcal{E}& = f_{cmp}=(C_p,C_v), \\
    \mathcal{S}& = f_{cmp}=(C_s,P_b),
\end{aligned}
\end{equation}
where $f_{cmp}$ denotes the Compare-Net\cite{shen2018improved}. To measure the embedding closeness and relevance, we design the comparison function as:
\begin{equation}
    f_{cmp}(C_1, C_2)=W_c[C_1, C_2, C_1-C_2,C_1*C_2],
\end{equation}
where $W_c$ is a transformation matrix and $*$ is Hadamard product. $C_1$ and $C_2$ are the features to be compared.
Additionally, we compare the news text with the results obtained from reverse search to verify the presence of temporal alignments. In particular, we concurrently splice temporal features in the vectors of the Compare-Net.
\begin{equation}
\begin{aligned}
    \mathcal{T} &= W_tT_g, \\
    \mathcal{R} &= f_{cmp}(C_r,P_b,\mathcal{T}) \\ 
    &=W_r[C_r,P_b,C_r-P_b,C_r*P_b,\mathcal{T}],
\end{aligned}
\end{equation}
where $\mathcal{R}$ represents the temporal consistency features, $W_t$ is is a 1-dimensional learnable matrix, $W_r$ refer to learnable transformation matrix.

\subsection{Training and Inference}

To obtain a better fake news representation of various attributions, we incorporate a classification head before each category of features to perform a binary classification task for distinguishing between real and fake news. The label for this task is denoted as $y_b$. In particular, we also separately perform a binary classification task on the images feature $V_c$ to better distinguish samples of visual effectiveness.
We use binary cross-entropy loss to individually optimize these five feature categories:
\begin{equation}
\begin{aligned}
     \hat{y}_n&=MLP({n}),\ n=\mathcal{E},\mathcal{S},\mathcal{R}, V_m, V_c,   \\
     \mathcal{L}_{n} &= -(y_{b}\cdot\operatorname{log}\hat{y}_{n}+(1-y_{b})\cdot\operatorname{log}(1-\hat{y}_{n})).
\end{aligned}
\end{equation}

Simultaneously, we concatenate the features and multiply the probability $\phi_n$ of a single judgment network indicating the news as fake with the corresponding network's feature. When the probability approaches 1, it signifies a higher likelihood of the news being false due to that particular feature. Meanwhile, we splice text clip semantic features to better obtain a global multimodal representation of the news. After passing through a Multilayer Perceptron (MLP), we obtain the final prediction result $\hat{y}_b$, \ie,
\begin{equation}
\begin{aligned}
    \hat{y}_b=MLP([P_c&,\mathcal{E}*\phi_\mathcal{E},\mathcal{S}*\phi_\mathcal{S},\\
\mathcal{R}*\phi_\mathcal{R}&,{V}_m*\phi_{m},{V}_c*\phi_{c}]).
\end{aligned}
\end{equation}

Then, we consider the minimization of the standard binary cross-entropy loss value as the objective function,\ie,
\begin{equation}
\begin{aligned}
     \mathcal{L}_b(y_b, \hat{y}_b)=-(y_b \log \hat{y}_b  &+ (1-y_b) \log (1-\hat{y}_b)) \\ 
     &+\frac{1}{5}\sum_n{\mathcal{L}_n},
 \end{aligned}
\end{equation}
where $y_b$ denotes the actual label and $y_b \in \{0,1\}$; $\hat{y}_b $ represent the predicted label.
In attributing inference, we define the downstream task as a 6-classification task, and obtain the final attributing prediction $\hat{y}$ through the MLP,\ie,
\begin{equation}
\begin{aligned}
\hat{y}=MLP([P_c&,\mathcal{E}*\phi_\mathcal{E},\mathcal{S}*\phi_\mathcal{S}, \\ \mathcal{R}*\phi_\mathcal{R}&, {V}_m*\phi_{m},{V}_c*\phi_{c}]),
\end{aligned}
\end{equation}
and optimize the classification results using cross-entropy loss,\ie,
\begin{equation}
    \mathcal{L}=-\sum_{i=1}^6y_i\log\hat{y}_i+\frac{1}{5}\sum_n{\mathcal{L}_n},
\end{equation}
where $\hat{y}_i$ is the probability of predicting the sample as class $i$.

\section{Experiment}
\label{sec:experiments}


\myparagraph{Experimental Settings}
Experimental settings can be found in Supplementary, which includes compared methods, implementation details, and evaluation metrics.
All experiments are conducted on a cluster of 8 RTX3090 GPUs. 
Additionally, we also analyze the \textbf{computational complexity} of the model; details can be found in the Supplementary.

\myparagraph{Results on Multimodal Fake News Detection}
According to Table~\ref{experiment1}, our proposed model exhibits the best performance across various metrics. \our achieves an approximately 2.5\% improvement in overall accuracy (acc) and a 2.8\% improvement in F1 score.
Additionally, to demonstrate the \textbf{generalization} of \our, we conduct experiments on public datasets Twitter~\cite{65boididou2015verifying}, Weibo~\cite{jin2017multimodal}, and Weibo21~\cite{nan2021mdfend}. \our outperforms the compared baselines, achieving F1-scores of 0.905, 0.899, and 0.901, respectively. Related table can be found in Supplementary.

\myparagraph{Discussion on Dataset Difficulty}
Comparing the experimental results of the same model on previous datasets, we observe that \amg is a more challenging dataset. BMR achieves an accuracy (acc) of 90\% on both the Weibo~\cite{jin2017multimodal} and GossipCop~\cite{64shu2020fakenewsnet}, while the detection accuracy of \amg falls below 81\%. Other models also exhibit varying degrees of performance decline. 
We have analyzed the reasons behind the increased challenge in \amg and arrived at a preliminary conclusion: the presence of \textbf{entity bias}~\cite{zhu2022generalizing} in the collection process of real and fake news. However, our approach of collecting true news has successfully avoided this bias.

\myparagraph{Results on Multimodal Fake News Attribution}  
We present the overall attribution accuracy and F1 scores in Table~\ref{experiment1}, while the detailed results on each attribution category are presented in Supplementary. The experimental results show that our model outperforms the baseline model in terms of overall attribution accuracy and F1 scores.
Compared to the suboptimal model BMR, our model achieves improvements of approximately 7\% and 4.7\% in accuracy and F1 score, respectively. Furthermore, \our demonstrates a significant enhancement of around 10\% in accuracy compared to other methods.

\begin{table}[tp]
\centering
\caption{Results of multimodal fake news detection and attribution.}
\label{experiment1}
\scalebox{0.92}{
\begin{tabular}{lcccc}
\Xhline{1.5px}
\hline
\multicolumn{1}{c}{\multirow{2}{*}{Method}} & \multicolumn{2}{c}{Fake News Detection} & \multicolumn{2}{c}{Fake News Attribution} \\ \cline{2-5} 
\multicolumn{1}{c}{}                        & Accuracy           & F1 Score           & Accuracy            & F1 Score            \\ \hline
CLIP                                        & 0.7812             & 0.7809             & 0.6469              & 0.5325              \\
CAFE                                        & 0.7667             & 0.7628             & 0.6382              & 0.4665               \\
MCAN                                        & 0.7740             & 0.7693             & 0.6115              & 0.4605              \\
BMR                                         & 0.8079             & 0.8057             & 0.6687              & 0.5193              \\ \hline
\textbf{\our}                               & \textbf{0.8323}    & \textbf{0.8310}    & \textbf{0.7385}     & \textbf{0.5666}     \\  \Xhline{1.5px}
\end{tabular}}
\end{table}

\begin{table}
    \centering
    \caption{F1 results of ablation study.}
    \label{tab:ablation}
    \setlength{\tabcolsep}{12pt}
    \scalebox{0.82}{
    \begin{tabular}{l c c c c}
    \Xhline{1.5px}
        \multirow{2}*{Method} & \multicolumn{2}{c}{Detection}  & \multicolumn{2}{c}{Attribution} \\ \cline{2-5}
        ~           & Acc     & F1     & Acc    & F1  \\ \hline
        w/o PSCC-NET & 0.8167 & 0.8166 & 0.6781 & 0.4660 \\ 
        w/o entity & 0.8146 & 0.8138 & 0.6937 & 0.4283   \\ 
        w/o event & 0.7917 & 0.7916 & 0.6813 & 0.4542   \\ 
        w/o temporal & 0.8094 & 0.8085 & 0.7010 & 0.4294   \\ 
        w/o vem & 0.8187 & 0.8177 & 0.6937 & 0.4407 \\ \hline
        \textbf{\our} & \textbf{0.8323} & \textbf{0.8310} & \textbf{0.7385} & \textbf{0.5666} \\ \Xhline{1.5px}
    \end{tabular}}
\end{table}

    

\myparagraph{Ablation Study}
To demonstrate the effectiveness of the multi-granularity clue and various feature extraction modules we employed, we conducted ablation experiments. The results of these experiments are displayed in Table~\ref{tab:ablation}.

Removing individual modules leads to a certain degree of decline in both detection and attribution performance. Among them, the removal of event-level coherence features has the greatest impact on the detection of multimodal fake news, resulting in a decrease of approximately 4\% in both accuracy and F1 score. 
Furthermore, the temporal coherence, which is the focus of \our, also has a significant impact on both detection and attribution, demonstrating the importance the temporal information between image and text.

\begin{figure}[t] 
    \centering
     \includegraphics[width=1\linewidth]{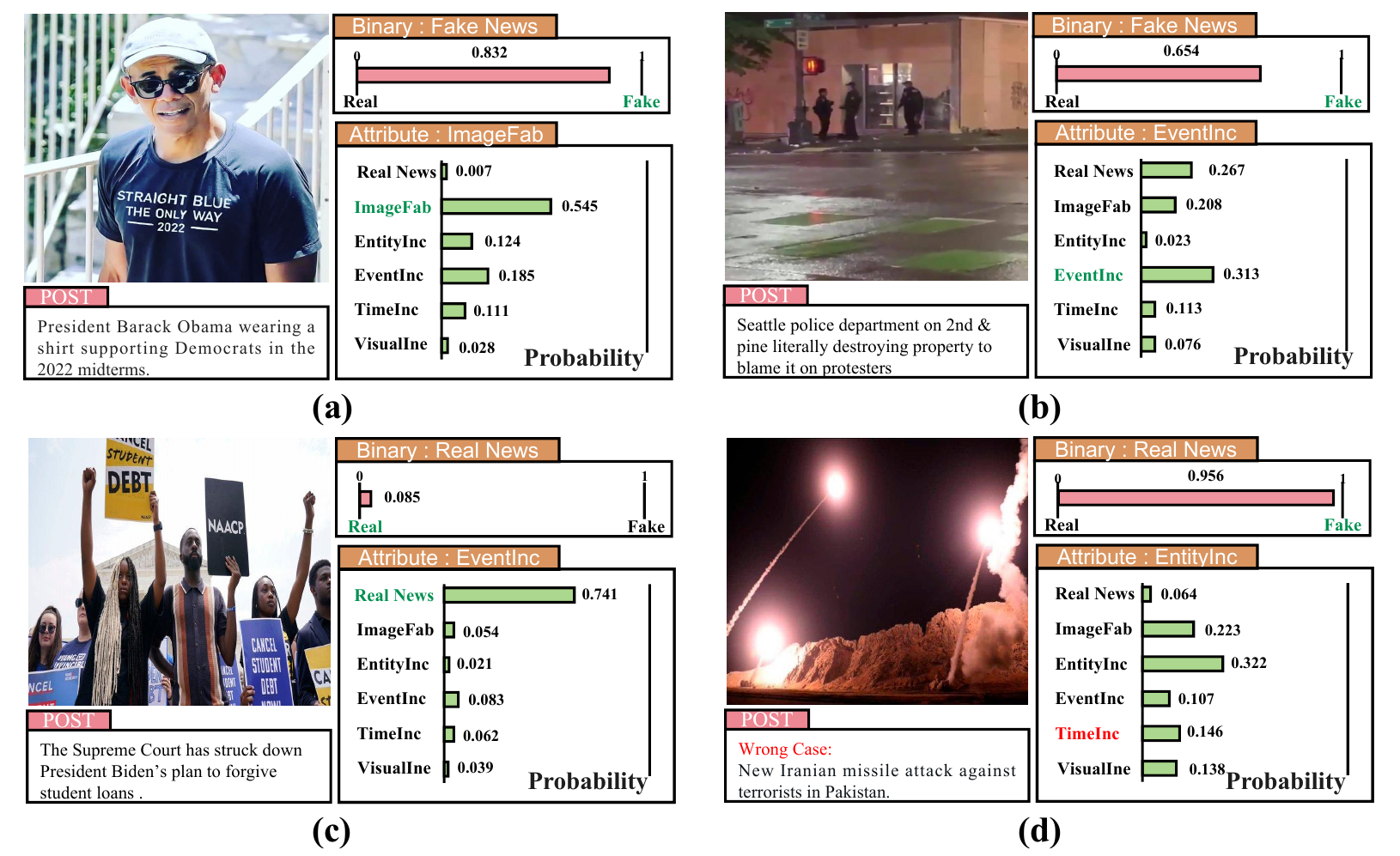}
    \caption{The case study of \our in \amg.}
    \label{cs}
\end{figure}

\myparagraph{Case Study}
We select four representative samples to analyze the detection and attribution results. As can be observed in Figure~\ref{cs}(a), (b) and (c), both detection and attribution produce accurate results.

For instance, in  Figure~\ref{cs}(a), despite the high coherence between the image and text, \our is still able to draw the conclusion of image fabrication.
However, there still exist some challenging news samples. As shown in Figure~\ref{cs}(d), this news claims that Iran used missiles to strike terrorists in Pakistan in 2018, but in reality, the image used in the article is from 2015, showcasing a typical case of time inconsistency. Although it is classified as fake news, it is categorized as entity inconsistency in the attribution process. 
 In the image, the key entity of "terrorists" mentioned in the text is not detected, which may leads the model to make the judgment.

\myparagraph{Discrimination Performance}
We utilize heatmaps to visualize the discriminative power of \our on \amg. We randomly select 90 real news and 90 fake news. We then calculate the pairwise similarities between the 16-dimensional representations from the binary classification classifier and the attribution classifier. 
The darker colors indicate weaker correlation and lighter colors indicate stronger correlation.

\begin{figure}[t] 
    \centering
     \includegraphics[width=1\linewidth]{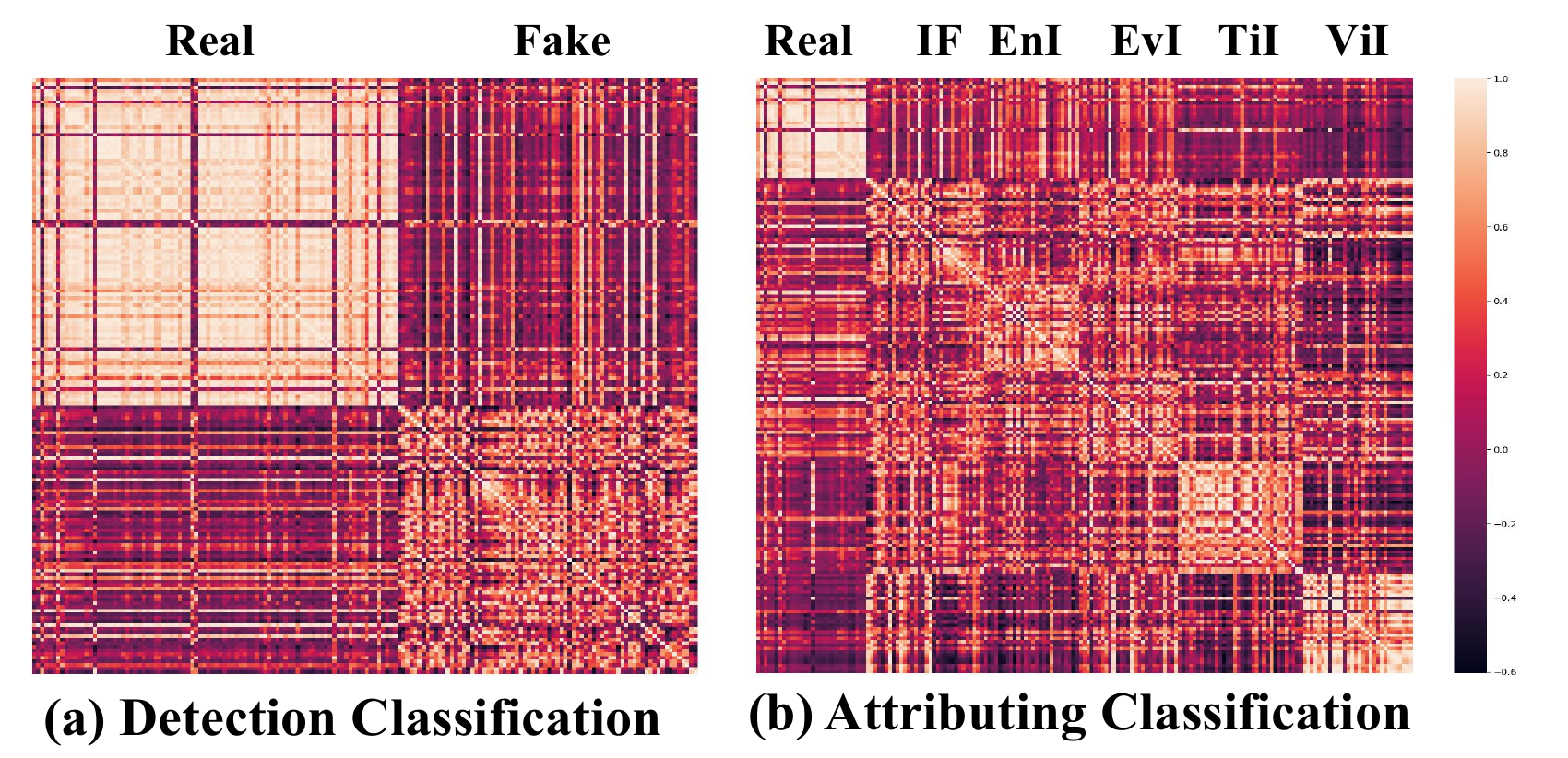}
    \caption{The discriminative power of \our.}
    \label{54}
\end{figure}

From Figure~\ref{54}, we can observe that our model demonstrates strong discriminative ability, with relatively clear intra-class similarity and inter-class differences. Additionally, it is evident that the binary classification representations of genuine news and fake news exhibit a higher level of distinctiveness, while the attribution learning shows a slightly reduced discriminative capacity. This observation indicates that capturing intra-class variations among the fake news instances represents the main challenge faced by \amg.

\section{Conclusion}
\label{sec:conclusion}
In this study, we introduce a novel task, multimodal fake news attribution, which aims to enhance the credibility of model detection results. We believe it will provide promising and meaningful avenues for research. Furthermore, we develop \amg, the first multimodal fake news attribution dataset and make it open-sourced, which will facilitate future follow-up studies. We emphasize the significance of temporal information in the detection of multimodal fake information, highlighting it as a key factor for fake news detection. We also introduce a competitive method \our.

\begin{figure}[t] 
    \centering
     \includegraphics[width=1\linewidth]{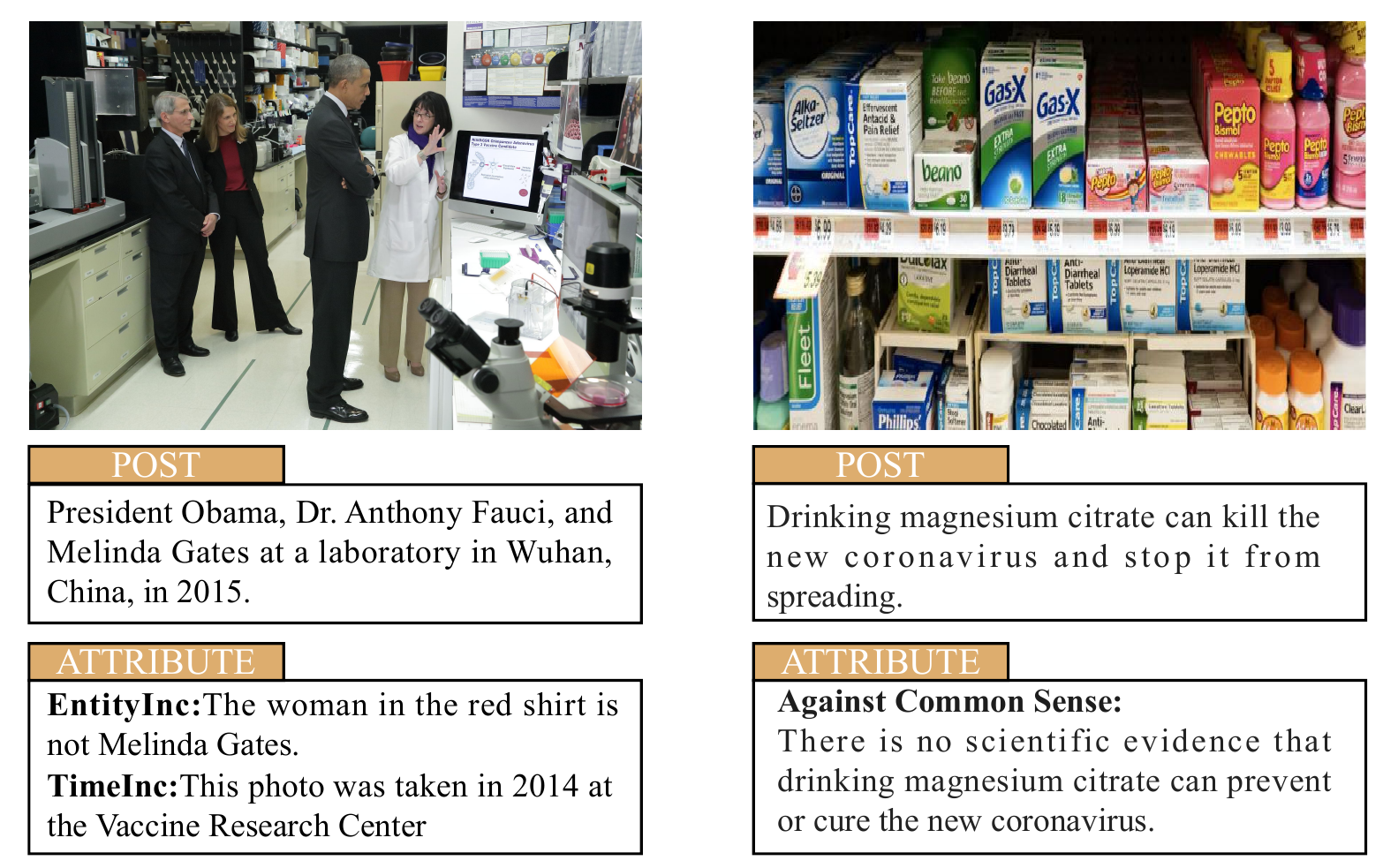}
    \caption{Fake News out of our attributions.}
    \label{outofatt}
\end{figure}

\myparagraph{Limitation}
\amg focuses solely on the content of multimodal fake news, excluding metadata like comments and social networks. We are collecting this data for future release. Additionally, besides dividing attributions into five categories, we include the label ``Not fall into any of the above types'', during the labeling process. 
Figure~\ref{outofatt} illustrates several instances that fall outside the scope of our attributions. In this regard, (a) delineates the occurrence of multiple overlapping attribution anomalies, encompassing both entity and temporal inconsistency. On the other hand, (b) signifies instances that do not conform to any of our attribution categories.


\section*{Acknowledgments}
This work was partially supported by National Key R\&D Program of China No. 2022YFB3102600, NSFC under grant Nos. U23A20296, 62272469, 62302513, 62192781 and 62272374.



\bigskip

\bibliography{aaai25}
\appendix
\newpage
\clearpage

\section{Appendix}
\subsection{Additions to Annotation Process}
\label{addannotation}
\begin{figure}[t] 
    \centering
     \includegraphics[width=0.9\linewidth]{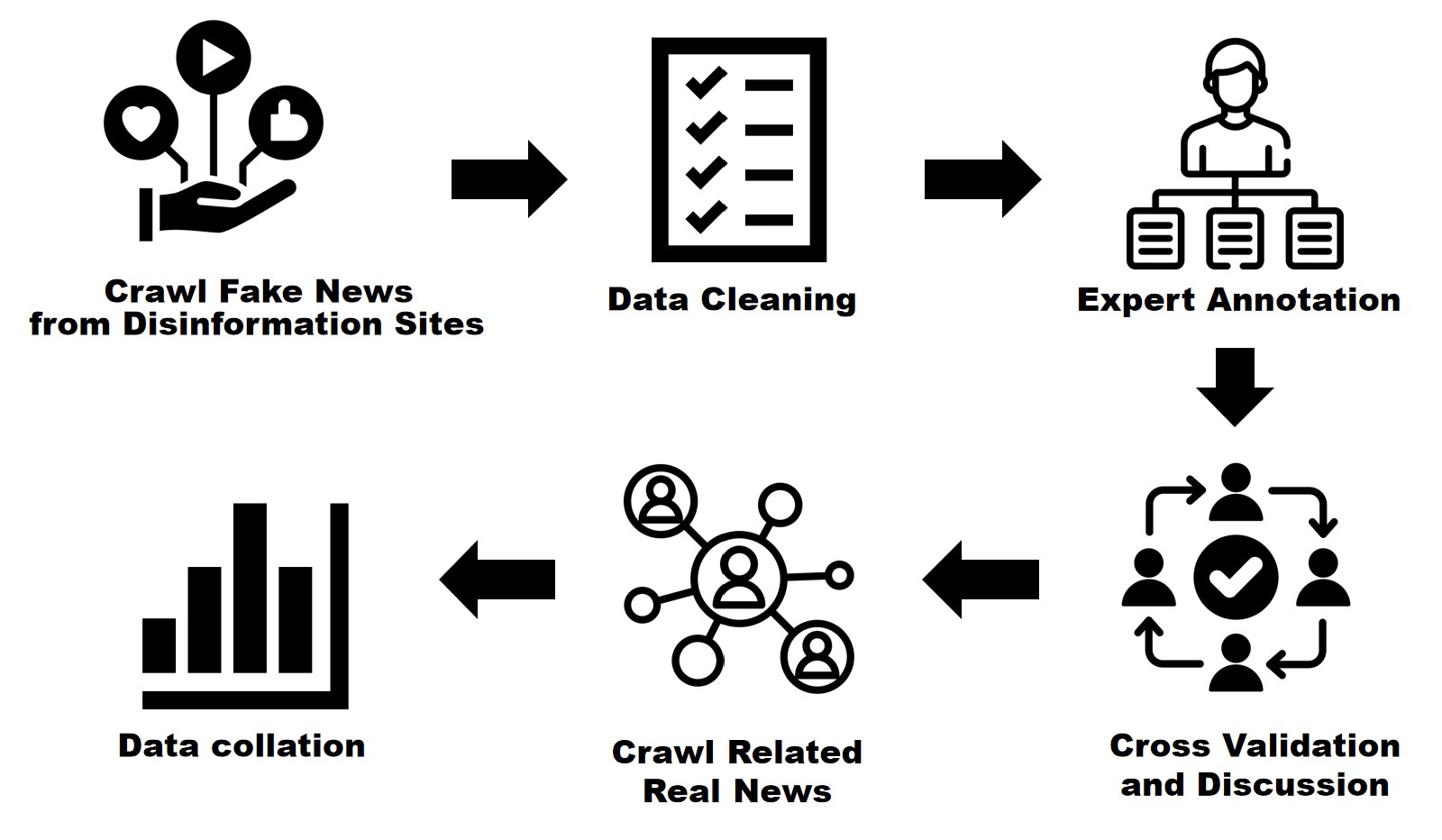}
    \caption{Process of dataset construction.}
    \label{dataprocess}
    \vspace{-0.1cm}
\end{figure}

Figure\ref{dataprocess} illustrates our annotation process, as described in Section 3.2, we went through six steps: data crawling, data cleaning, expert annotation, cross-validation and discussion, related real news crawling, and data collation. In the process of expert annotation, in order to obtain accurate and robust annotations, we invited 17 researchers in the fields of computer science and news communication to participate in the annotation, who are familiar with the research on the dissemination and detection of fake news, and most of them have published related articles or designed detection systems. 
All annotators underwent rigorous training and were well-versed in data privacy and security regulations.
Specifically, we carefully selected 100 typical cases through group and unified discussions, and assigned each news item to three experts. We then asked them to attribute each sample, with the option of choosing between "does not belong to any category" and "not sure".
We calculated the accuracy and F1 scores between the expert labeling and the labels of the pre-selected typical dataset before voting, and only allowed the expert labeled results to go to the voting stage when all the metrics were above 95\%.

\subsection{Examples of Various Category }
\label{examples}


\subsection{More Statistics of \amg}
\label{moresta}
\begin{figure}[t] 
    \centering
     \includegraphics[width=0.8\linewidth]{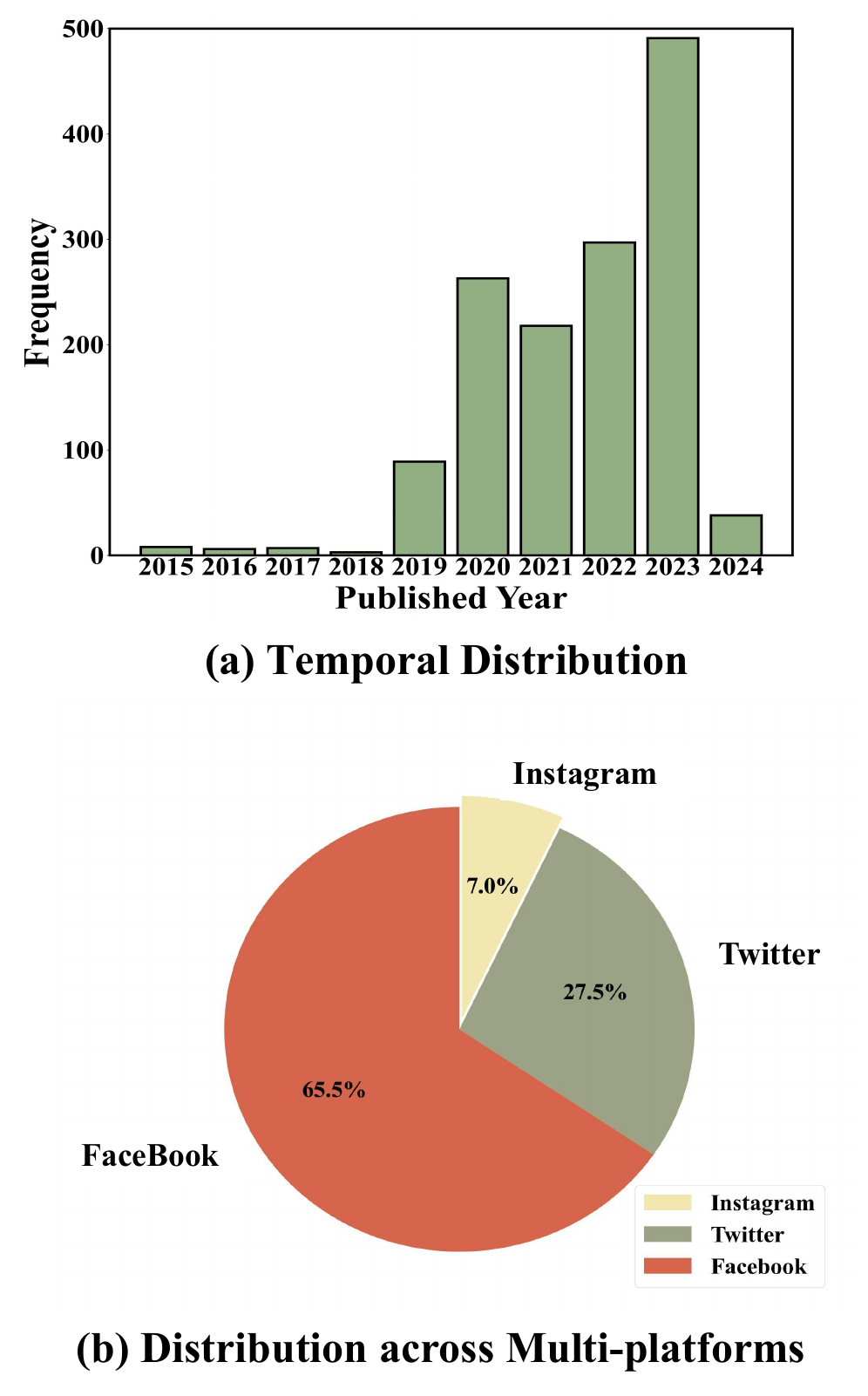}
     \vspace{-0.1cm}
    \caption{The characteristics of \amg.}
    \label{timed}
\end{figure}

We summarize the statistics of temporal distribution and multi-platforms distribution in
Figure~\ref{timed} in \our.

\subsection{Preliminary}
\label{preliminary}

\myparagraph{Task Definition}
\amg contains both binary labels for real and fake classification, as well as multi-class labels for attributing different types of errors. Consequently, we conduct two tasks: multimodal fake news detection and multimodal fake news attribution. A piece of multimodal news can be represented as$\{(p,v),y_b\in\{0,1\} ,y\in\{0,1,2,3,4,5\}\}$, where $p$ and $v$ represent the textual and visual content, respectively. And $y_b$ denotes the detection label and and $y$ is attribution label.

\noindent \textbf{Task1. Multimodal Fake News Detection:}
Given a piece of multimodal news, it seeks to categorize news pieces into fake or real.
Each piece of news contains the textual and visual contents, and has a ground-truth label $y_b \in \{0,1\}$, such that 1 denotes fake, 0 denotes real. 

\noindent \textbf{Task2. Multimodal Fake News Attribution:}
Given a piece of multimodal news, attributting task aims to determine the authenticity of news while attributing the reasons behind its falsehood to five pre-defined categories.
Each piece of fake news has a ground-truth label $y \in \{0,1,2,3,4,5\}$, which represents real news, ImageFab, ImageNoE, EntityInc, EventInc, or TimeInc, respectively.

\myparagraph{Video Preprocessing} In the case of video in visual modality, we conduct preliminary processing. We consider the middle frame of the video to be crucial. Therefore, we extract the middle frame as a representative for video. Therefore, in the subsequent model section, our treatment of the visual modality specifically refers to images. Exploring improved methods for handling the video modality is an area of future research for us.

\subsection{Experiment Setting}
\label{expset}

In this section, we briefly introduce compared approaches, implementation details and evaluation metrics.

\myparagraph{Baselines}
We compare our proposal with \sota solutions, including:
\begin{itemize}
\item MCAN~\cite{wu2021multimodalmcan} proposes multimodal Co-Attention
Networks to better fuse textual and visual features for fake news detection.
\item CAFE~\cite{chen2022cross} improves fake news detection accuracy by adaptively aggregating unimodal features and cross-modal correlations.
\item BMR~\cite{ying2023bootstrapping} presents a novel scheme of bootstrapping multi-view representations for fake news detection. It extracts the views of the text, the image pattern and the image semantics, then proposes iMMoE for feature refinement and fusion.
\item CLIP~\cite{radford2021learning} is used to normalize the representation of image and text, and then a joint representation is used for prediction.

\end{itemize}

\myparagraph{Settings of LLaVA}
We utilized the pre-trained multimodal large language model LLaVA-1.5~\cite{liu2023llava,liu2023improvedllava} for event extraction from images, employing 4-bit quantization. The prompt for this task was set as: ``describe the event of the image breifly''.

\begin{table}
    \centering
    \caption{Results of multimodal fake news attribution.}
    \label{experiment2-2}
    \setlength{\tabcolsep}{10pt}
    \scalebox{0.7}{
    \begin{tabular}{ l c c c c c c}
    \Xhline{1.5px}
        Method & acc & F1 & att. & precision & recall & F1   \\ \hline
        \multirow{6}*{CAFE} &  \multirow{6}*{0.6382} &  \multirow{6}*{0.4665} & 0 & 0.8021 & 0.8354 & 0.8184  \\ 
        ~ & ~ & ~ & 1 & 0.3108 & 0.2212 & 0.2584   \\ 
        ~ & ~ & ~ & 2 & 0.6957 & 0.7273 & 0.7111   \\
        ~ & ~ & ~ & 3 & 0.3134 & 0.4118 & 0.3559   \\ 
        ~ & ~ & ~ & 4 & 0.4176 & 0.3065 & 0.3535   \\ 
        ~ & ~ & ~ & 5 & 0.2759 & 0.3333 & 0.3019   \\
        \hline
        \multirow{6}*{MCAN} & \multirow{6}*{0.6115} & \multirow{6}*{0.4605} & 0 & 0.8102 & 0.7544 & 0.7813 \\ 
        ~ & ~ & ~ & 1 & 0.2000 & 0.1757 & 0.1871   \\ 
        ~ & ~ & ~ & 2 & 0.5625 & 0.7941 & 0.6585   \\
        ~ & ~ & ~ & 3 & 0.3500 & 0.3158 & 0.3320   \\ 
        ~ & ~ & ~ & 4 & 0.3071 & 0.4333 & 0.3594   \\ 
        ~ & ~ & ~ & 5 & 0.4800 & 0.4138 & 0.4444   \\ 
        \hline
         \multirow{6}*{BMR} & \multirow{6}*{0.6687} & \multirow{6}*{0.5193} & 0 & 0.8998 & 0.8135 & 0.8545 \\ 
        ~ & ~ & ~ & 1 & 0.2021 & 0.2568 & 0.2262   \\ 
        ~ & ~ & ~ & 2 & 0.5775 & 0.5942 & 0.5857   \\
        ~ & ~ & ~ & 3 & 0.3838 & 0.5299 & 0.4451   \\ 
        ~ & ~ & ~ & 4 & 0.4878 & 0.4396 & 0.4624   \\ 
        ~ & ~ & ~ & 5 & 0.6842 & 0.4483 & 0.5417   \\
         \hline
        \multirow{6}*{CLIP} & \multirow{6}*{0.6469} & \multirow{6}*{0.5325} & 0 &  0.9196 & 0.7215 & 0.8086 \\
        ~ & ~ & ~ & 1 &  0.2500 & 0.3889 & 0.3043   \\ 
        ~ & ~ & ~ & 2 &  0.6377 & 0.6567 & 0.6471   \\
        ~ & ~ & ~ & 3 &  0.3764 & 0.5076 & 0.4323  \\ 
        ~ & ~ & ~ & 4 &  0.4370 & 0.6556 & 0.5224   \\ 
        ~ & ~ & ~ & 5 &  0.6111 & 0.3929 & 0.4783  \\
        \hline
        \multirow{6}*{\textbf{\our}} & \multirow{6}*{ \textbf{0.7385}} & \multirow{6}*{\textbf{0.5666}} & 0 &  0.9293 & 0.8953 & 0.912 \\
        ~ & ~ & ~ & 1 &  0.3544 & 0.4000 & 0.3758   \\ 
        ~ & ~ & ~ & 2 & 0.6234 & 0.7273 & 0.6713  \\ 
        ~ & ~ & ~ & 3 & 0.4815 & 0.3881 & 0.4298   \\ 
        ~ & ~ & ~ & 4 & 0.459 & 0.6292 & 0.5308    \\ 
        ~ & ~ & ~ & 5 &  0.5455 & 0.4286 & 0.4800   \\
        \Xhline{1.5px}
    \end{tabular}}
\end{table}

\myparagraph{Implementation Details}
We employ the CLIP ViT-B/16 model for image-text feature extraction, while the bert-base-cased model is utilized for processing the multiple clue extracted from image and text.
Both BERT and CLIP models are kept frozen, ensuring their pre-trained weights are preserved. 
All MLP (Multi-Layer Perceptron) layers consist of a hidden layer with 256 dimensions, followed by Batch Normalization (BatchNorm 1D), and ReLU activation function. 
For optimization, we use the Adam optimizer. The batch size is set to 64, and the learning rate is $1\times10^{-4}$. 
Image sizes are adjusted to 224×224 for consistency across experiments. 
All experiments are conducted on a cluster of 8 RTX3090 GPUs.
\vspace{-0.2cm}

\subsubsection{Evaluation Metric.} 
We employ accuracy(acc) as the primary evaluation metric for multimodal fake news detection and attribution. Considering the imbalanced nature of label distribution, we additionally incorporate precision, recall, and F1 score as complementary evaluation metrics alongside accuracy.


\subsection{Detailed Results of Multimodal Fake News Attribution}
\label{expatt}
Table~\ref{experiment2-2} shows the results for each attribution category.
We further analyze the attribution results for each category. The accuracy for image fabrication is relatively low. This could be attributed to the fact that image forgery techniques encompass various types, and PSCC-NET can only model one of them. 
The detection accuracy for entity inconsistency does not meet our expectations. We find that APIs used for image entity recognition have relatively poor accuracy, which hinders the effectiveness of this type of detection.

\subsection{Model Generalization}
Our proposed \our aims to solve complex attribution problems by targeting typical errors in current fake news. By leveraging external evidence, \our is expected to improve performance on classic fake news detection datasets. We conduct experiments on public datasets Twitter~\cite{65boididou2015verifying}, Weibo~\cite{jin2017multimodal}, and Weibo21~\cite{nan2021mdfend}. 

As shown in Table~\ref{tab:MG}, \our achieves the best results across all datasets. On the Twitter dataset, it outperforms CLIP by over 2\%, and on the Chinese datasets, its F1-score was approximately 1\% higher than the next best model.

\begin{table}
\caption{F1 score of \our in other datasets.}
\label{tab:MG}
\setlength{\tabcolsep}{11pt}
\scalebox{0.85}{
\begin{tabular}{l c c c c c}
\hline
Method & CAFE & MCAN & BMR & CLIP & \our\\
\hline
Twitter & 0.869  & 0.875 & 0.872 & 0.883 & 0.905\\
Weibo &  0.855  & 0.871  & 0.884   & 0.887 & 0.899\\ 
Weibo21 & 0.882 & 0.896  & 0.900 &  0.904 &  0.913\\
\hline
\end{tabular}}
\end{table}




\section{Related Works} 

In this section, we briefly review and discuss the related works on methods and datasets of multimodal fake news detection.
In contrast to fact-checking~\cite{guo2022survey,34vlachos2014fact,35vlachos2015identification} or claim verification~\cite{schlichtkrull2024averitec, jiang2020hover, wu2021evidence}, fake news detection~\cite{guo2023interpretable, 9802916m3fend, 6shu2017fake} does not provide additional ground truth textual or visual evidence, and the objective of detection is to verify the authenticity of posts sourced from social platforms, rather than conclusive claims. So methods~\cite{zhou2019gear,zhong2020reasoning} and datasets~\cite{34vlachos2014fact,yao2023end} related to fact-checking are not within the scope of our discussion.

\myparagraph{Weakness of Existing Datasets} Table~\ref{alldataset} presents the widely used datasets for multimodal fake news detection, and we summarize the following points:
1) Binary Label Scheme: Most datasets focus on binary classification (real or fake). Fakeddit~\cite{nakamura2020fakeddit} stands as the sole fine-grained multimodal dataset, whose classification is based on the amalgamation of various subreddits on the Reddit platform. The research on fine-grained annotation of attribution remains largely void.
2) Out-of-Date: Most mainstream datasets were collected before 2020 and the evolution of technology in recent years has given rise to new forms of multimodal fake news.
3) Single Platform Source: English datasets are primarily sourced from Twitter, overlooking other platforms, such as Facebook and Instagram. Facebook is particularly notorious for being a hotspot of fake news. 
4) Specific domain: Some datasets are tailored to specific domains or events, casting doubt on the generalizability of models trained on such datasets, including Pheme, Politifact, and ReCOVery.

\myparagraph{Multimodal Fake News Detection} Several multimodal fake news detection methods primarily focus on designing models that combine textual and visual features to determine authenticity~\cite{jin2017multimodal,wu2021multimodalmcan,wu2023see,wang2023cross}. 
Some studies incorporate the cross-modal correlations between images and text in the detection framework~\cite{safe, xue2021detectingMCNN, chen2022cross,hsen23}.
Some methods take the frequency domain feature~\cite{wu2021multimodalmcan, xue2021detectingMCNN} and the pixel domain feature~\cite{qi2019exploitingmvnn,59jin2016novel} of images into consideration, reflecting digital alterations within the images. 
External knowledge graphs~\cite{wang2020fake, zhang2024reinforced} and  crowd wisdom like comments~\cite{cui2019same, wu2023see} are introduced to facilitate fake news detection.
Several approaches based on logical reasoning~\cite{liu2023interpretable}, neuro-symbolic reasoning~\cite{Dong_He_Wang_Jin_Ge_Yang_Jin_2024} and causal intervention~\cite{chen2023causal} have been proposed for improving interpretability of detection process. 
Additionally, short videos have become a popular channel for news dissemination, prompting recent research into detecting fake news in video formats~\cite{fakesv, TikTec, need}.

\end{document}